\documentclass[preprint,sort]{elsarticle}
\usepackage{graphicx}
\usepackage[noend,linesnumbered,figure,boxed]{algorithm2e}

\begin{document}
\begin{frontmatter}

\title{Shape Reconstruction and Recognition with Isolated Non-directional Cues}
\author[Math]{Toshiro Kubota\corref{cor1}}
\ead{kubota@susqu.edu}
\author[Biol]{Jessica Ranck}
\ead{ranck@susqu.edu}
\author[Math]{Briley Acker}
\ead{ackerb@susqu.edu}
\author[Math]{Herman De Haan}
\ead{dehaan@susqu.edu}

\cortext[cor1]{Corresponding author}
\address[Math]{Department of Mathematical Sciences
\\Susquehanna University\\Selinsgrove PA 18970, USA}
\address[Biol]{Department of Biology\\
Susquehanna University\\Selinsgrove PA 18970, USA}

\begin{abstract}
The paper investigates a hypothesis that our visual system
groups visual cues based on how they form a surface, or more
specifically triangulation derived from the visual cues. To
test our hypothesis, we compare shape recognition with three
different representations of visual cues: a set of isolated
dots delineating the outline of the shape, a set of triangles
obtained from Delaunay triangulation of the set of dots, and a
subset of Delaunay triangles excluding those outside of the
shape. Each participant was assigned to one particular
representation type and increased the number of dots (and
consequentially triangles) until the underlying shape could be
identified. We compare the average number of dots needed for
identification among three types of representations. Our
hypothesis predicts that the results from the three
representations will be similar. However, they show
statistically significant differences. The paper also presents
triangulation based algorithms for reconstruction and
recognition of a shape from a set of isolated dots. Experiments
showed that the algorithms were more effective and perceptually
agreeable than similar contour based ones. From these
experiments, we conclude that triangulation does affect our
shape recognition. However, the surface based approach presents
a number of computational advantages over the contour based one
and should be studied further.
\end{abstract}
\end{frontmatter}

\section{Introduction}\label{intro}
Grouping of visual cues is believed to be a fundamental step in
our visual perception\cite{Attneave:Shape,ullman:vision}. In
particular, great amount of attention has been paid to
fragments of object contours or edges as the visual cues
\cite{canny:edge,marr:vision}. Although primitive, edges
provide orientation information that can aid the grouping
process. Furthermore, tangential and curvature information can
be derived from a collection of edges via sophisticated
interpolation schemes
\cite{elder:curve,Sharon:PAMI00,Kimia:IJCV2003}. Similarly,
motion fields (optical flows) and surface normal are common for
similar goals in motion data\cite{Aggarwal:1988,Horn:1993} and
three-dimensional data\cite{Hoppe:1992,Frankot:1988},
respectively. They provide orientation information in a higher
dimension.

Despite significant amounts of efforts, no consensus has
emerged as to how edges can be grouped in our visual system.
The main problem is that grouping is hard. Given $N$ edges,
there are in the order of $N!$ possible ways to group them. By
incorporating information such as location, orientation, and
curvature, we can reduce the size of the solution space. But
the space appears to be still too large for any known
computational algorithms to search it within a reasonable
amount of time.

Another issue is that we often perform grouping of visual cues
effectively when cues do not possess any orientation
information. For example, if the outline of a shape is sampled
and these points are shown by small dots, we can often
recognize the original shape. It applies to other geometrical
primitives such as squares and crosses
\cite{julesz1981textons,bergen1988early}. These observations
have led some researchers questioning the edge grouping
approach \cite{Feldman:2001,Greene:PMS2007,Ommer:IJCV2009}.
Tangential and curvature information does help if available
\cite{Kovacs:Gestalt,field:neuron}. However, most proposals may
be too dependent on such information.

\begin{figure*}
\centering
\includegraphics[width=4.0in]{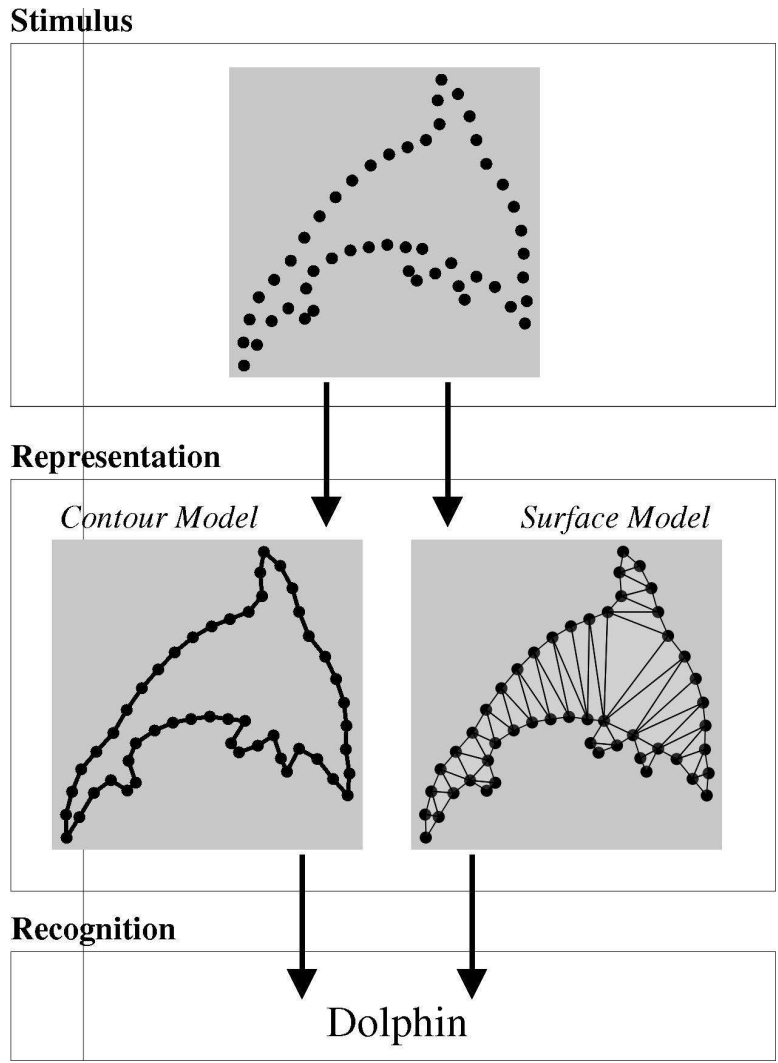}
\caption{Surface model vs. contour model. The Top image is a
visual stimulus shown to a subject, and the two middle pictures
represent the two internal models that may be constructed
internally to induce the recognition.} \label{fig:dolphin}
\end{figure*}

The hypothesis we formulated for the question is that when a
set of cues is presented, we tried to form a surface that fits
the set of cues. Directional information can help the process
but is not necessary. Without it, we are still capable of
performing the task of grouping non-directional cues. Figure
\ref{fig:dolphin} illustrates the two views: contour based and
surface based. The contour model is a more conventional
approach to the grouping problem, and the surface model is our
hypothesized approach. Given a well-organized set of primitives
such as the one shown at the top, we can no doubt recognize the
object as indicated at the bottom. In this case, the underlying
shape is a dolphin. The exact computational steps to reach from
the cues to the recognition are largely unknown. Nevertheless,
it is safe to assume that we construct some internal
representation, as we often experience such grouping
precepts\cite{Humphrey:Beauty,Koenderink:1979}. The contour
model considers a sequence of cues as the representation, while
the surface model considers a surface (height fields) derived
from the cues. In a sense, the surface model is more general
than the contour model, as we can always derive a contour
representation as a level set of the surface. On the other
hand, we cannot derive a surface from a contour unless the
contour forms a closed cycle. In such case, a surface can be
derived from the contour as a subsequent fill-in process
\cite{Elder:VR1998,Pessoa:BBS1998}.

Computationally, the two models shown in Figure
\ref{fig:dolphin} face different issues. For the contour model,
the issue is to find a permutation of cues, with which a
complete contour can be drawn. This is the contour integration
problem. For the contour model to be a feasible one to account
for the path from cues to recognition, an efficient and
reliable contour integration method needs to exist, and a large
number of algorithms have been proposed for the problem
\cite{Arbelaez:PAMI2011,cox:edgegrp,Grossberg:group,guy:percgrp,Li:Contour,Mahamud:PO,Papari:TIP2008,parent:curve,Saund:PO,shashua:sal,Wang:RatioContour}.
However, none has been shown satisfactory to account for the
performance of the human perception. Furthermore, as stated
above, most algorithms require directional information embedded
in the cues, and cannot handle cases like the one shown in
Figure \ref{fig:dolphin}.

For the surface model, we can fit a surface from a set of dots
by using a standard triangulation algorithm such as Delaunay
triangulation \cite{ComputationalGeometryBook}. Then, the issue
is to differentiate those triangles inside the boundary from
those outside. For such surface model to be a feasible one, an
efficient and reliable triangle differentiation method needs to
exist. We conjecture that such method exists, thus the internal
representation consists of two phases: straight triangulation
of cues, and triangulation after the differentiation.

To study plausibility of our hypothesis, we conduct two
studies: cognitive one and computational one. For the cognitive
study, we divide subjects into three different cohorts. One was
presented a set of dots representing a contour of a shape and
the other two were presented a set of triangles representing
the surface of the shape. The dot representation is the
simplest representation of the underlying shape. According to
our hypothetical model, the dot representation is converted
internally to a surface representation. Furthermore, the
surface representation evolves in two steps: all triangles
resulted from triangulation of dots and triangles that are
inside the underlying shape. Two triangle based representations
provide the two respective internal representations. If our
hypothetical model is correct, we expect to observe similar
recognition performance in all three cohorts.

For the computational study, we design a simple algorithm that
takes triangulation of dots and separates outer triangles from
inner ones. This process will bring grouping of visual cues as
detailed in Method Section. We also design another algorithm
that uses the grouping result to retrieve the shape from the
database of shapes. This retrieval procedure allows us to
compare the performance of the computational approach with the
above cognitive study. If an algorithm using the surface based
model outperforms a contour based algorithm of similar
computational complexity, the evidence favors the surface based
hypothesis over the contour based one. Furthermore, if the
performance of the surface based algorithm is comparable to the
human performance, the evidence favors the surface based model.

In the followings, we use bold capitals (\textbf{P}) for sets,
bold letters (\textbf{x}) for 2D structures such as points and
triangles, Greek letters ($\phi$, $\Omega$) for functions and
transformations, and italic alphabets ($K$, $n$) for scalar
variables. We use \textit{shape} to mean the boundary of an
object. We only consider shapes that are connected and have no
holes. Thus, a shape forms a connected simple cycle (i.e. a
connected cycle without self-intersection).

\section{Methods}\label{methods}
\subsection{Experiment 1 (Cognitive Study)}
Twenty subjects (10 male and 10 female) from Susquehanna
University, both undergraduate students and staff members, with
normal or corrected to normal vision participated in the study.
The experiment complies with the Code of Ethics of the World
Medical Association (Declaration of Helsinki) for experiments
involving humans and was approved by the Institutional Review
Board of Susquehanna University prior to its implementation.

\begin{figure}
\centering
\includegraphics[width=4.0in]{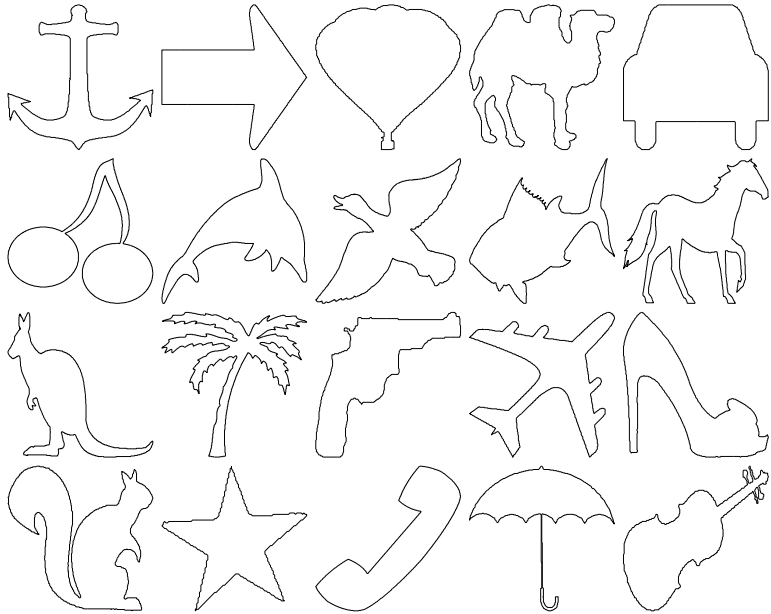}
\caption{Shapes used in our experiments.}
\label{fig:allShapes}
\end{figure}

The experiment was conducted in a computer lab lit by
fluorescent lighting. Each subject was seated in front of a
personal computer with a 17-inch monitor. All visual stimuli
were images with 512 by 512 pixels and displayed on a 5.6 by
5.6 $cm^2$ window in the monitor. Twenty shapes of animals and
objects shown in Figure \ref{fig:allShapes} were used in the
experiments. They were represented by a densely populated
sequence of points denoted by \textbf{S}. A computer program
written in Processing programming language took \textbf{S} and
generated visual stimulus. In addition to \textbf{S}, the
program required the number of sampled points ($K$).

\begin{figure}
\centering
\includegraphics[width=4.0in]{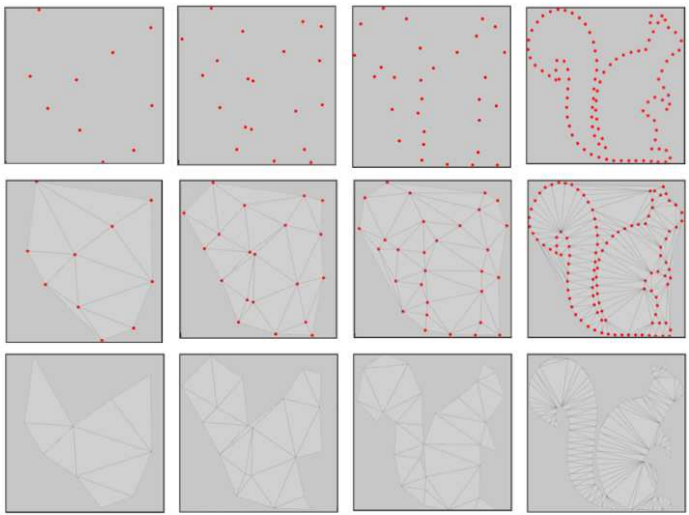}
\caption{Sample images of stimuli used in Experiment 1.
Each row shows a particular representation,
and each column shows images at different sample size ($K$).  Row
1: "Points"($\Omega_P$), Row 2: "All Triangles"($\Omega_A$),
Row 3: "Triangles"($\Omega_T$). Column 1: $K=10$, Column 2: $K=20$,
Column 3: $K=30$, Column 4: $K=100$.}
\label{fig:squirrel}
\end{figure}

The program can generate three types of representations: Points
representation, Triangle representation, and All-Triangle
representation. Let $\Omega_P$, $\Omega_T$, and $\Omega_A$
denote the Point, Triangle, and All-Triangle representations,
respectively. For $\Omega_P$, the program simply samples
\textbf{S} uniformly into $K$ sampled points denoted as
\textbf{P}, and then displays them on the window using red
disks of 10 pixel radius. For $\Omega_A$, the program first
prepares $\Omega_P$, and then runs Delaunay triangulation on
\textbf{P}. It then displays both points and triangles. Points
are shown with red disks of 10 pixel radius, faces of triangles
are shown in lighter shade of gray than the background, and the
sides of the triangles are shown in slightly darker shade of
gray than the background. For $\Omega_T$, the program first
prepares $\Omega_A$. For each triangle, the program checks if
the center of the triangle (the average of the three vertex
coordinates) is inside the shape. If not, the triangle is
removed. The program then displays only the remaining triangles
using the same shades used in $\Omega_A$. Figure
\ref{fig:squirrel} shows examples of $\Omega_P$, $\Omega_A$,
and $\Omega_T$ for a squirrel shape at $K=10$, 20, 30, and 100.

Among the twenty participants, 8 were shown $\Omega_P$, 6 were
shown $\Omega_A$, and 6 were shown $\Omega_T$. Each participant
went through all twenty shapes in a random order, but viewed
them only in the particular representation of the group. At the
beginning of viewing of a shape, $K$ is set to 10. Thus, the
$\Omega_P$ group saw 10 small disks at the 10 sample points.
The $\Omega_A$ group saw 10 disks and its Delaunay
triangulation. The $\Omega_T$ group saw a set of triangles in
the Delaunay triangulation whose centroids were inside the
shape delineated by the 10 sample points.

At $K=10$, most shapes are not recognizable. The participants
were allowed to increase $K$ at 10 increment by hitting the
SPACE bar. They were asked to increase $K$ until they were able
to recognize the shape shown by the particular representation.
At the time, they were asked to hit another key ('d'), which
brings a dialog with a list of shapes. They were asked to
select the recognized shape from the list. If no appropriate
shape was found in the list, the participants were able to go
back to the experiment and continue from the point where they
were left off. The participants were not able to reduce the
number of points. The program recorded both $K$ at the time of
recognition and the selected shape. The list in the dialog
remained the same throughout the experiment and provides names
of 20 shapes used in the experiment.

\subsection{Experiment 2 (Computational Study)}
In this section, we describe two dot grouping algorithms, one
surface based and the other contour based. We use them to
compare their performance in reconstructing the original shape.
We then describe a shape retrieval algorithm using the surface
based grouping. We use it to compare the algorithmic retrieval
against the results of the first experiment.

We first describe a grouping algorithm with the surface based
representation, which differentiates triangles inside a shape
and those outside the shape. We do this by exploring a planar
Hamiltonian cycle of the points from which the triangulation is
derived. The overall strategy is the following. Let
$\textbf{P}=[\textbf{p}_1, \textbf{p}_2, \cdots, \textbf{p}_K]$
be the set of points in the order that delineates the shape.
Some of the points are at the boundary of the triangulation,
which we can sequence. Let $\textbf{B}=[\textbf{b}_1,
\textbf{b}_2, \cdots, \textbf{b}_k]$ be the sequence of
boundary points. For simplicity and without loss of generality,
assume both $\textbf{P}$ and $\textbf{B}$ are given in the
clockwise order around the shape.

We can show that the order of points in \textbf{B} is preserved
in \textbf{P}. In other words, take a pair of adjacent points
in $\textbf{B}$ (with possible wrap around), say $\textbf{b}_i$
and $\textbf{b}_j$, and find the corresponding points in
\textbf{P}, say $\textbf{p}_{i'}=\textbf{b}_i$ and
$\textbf{p}_{j'}=\textbf{b}_j$. Then, no other points from
\textbf{B} can appear in the clockwise path from
$\textbf{p}_{i'}$ to $\textbf{p}_{j'}$ in \textbf{P}. This is
so, because a triangulated graph of triangulation is planar and
violation of the above property results in a non-planar
Hamiltonian path. The property allows us to refine \textbf{B}
by removing some outer triangles and exposing points that are
missing in \textbf{B}. When all points are exposed, we obtain a
planar Hamiltonian cycle of the point set. Of course, we do not
know the sequence in \textbf{P}, thus which triangle to remove.
However, observing Figure \ref{fig:squirrel}, it appears that
we can use some heuristics to differentiate those outside the
shape and those inside. More specifically, those triangles that
are outside the shape tend to be 'flat'. This trend becomes
more prominent as the sample size ($K$) increases since one
side of a triangle which connects adjacent pair of points in
the shape decreases while the other two which connect
non-adjacent pairs remain relatively unchanged.

What we need is a measure of flatness. Empirically, we found
the following effective. We denote a boundary edge connecting
points $\textbf{x}$ and $\textbf{y}$ as
$(\textbf{x},\textbf{y})$. Since a boundary edge is part of a
single triangle in the triangulation, we can uniquely associate
the edge with the other vertex of the triangle. Let the other
vertex denoted as $\textbf{xy}$. Then the flatness of
$(\textbf{x},\textbf{y})$ is defined as
\begin{equation}\label{eq:flatness}
\phi(\textbf{x},\textbf{y})=\|\textbf{x}-\textbf{y}\|
-\min(\|\textbf{x}-\textbf{y}\|,\|\textbf{x}-\textbf{xy}\|,\|\textbf{y}-\textbf{xy}\|)
\end{equation}
where $\|\textbf{x}-\textbf{y}\|$ is the length of the edge
$(\textbf{x},\textbf{y})$. In other words, the flatness of a
boundary edge $(\textbf{x},\textbf{y})$ is the length of the
edge minus the shortest side of the triangle associated with
$(\textbf{x},\textbf{y})$. Note that we can remove a boundary
edge $(\textbf{x},\textbf{y})$ only if $\textbf{xy}$ is not at
the boundary (or if it is \textit{internal}). Else, the trace
of the boundary points will not be a simple cycle. We call a
boundary edge $(\textbf{x},\textbf{y})$ \textit{removable}, if
$\textbf{xy}$ is internal.

\begin{figure}
\centering
\includegraphics[width=2.0in]{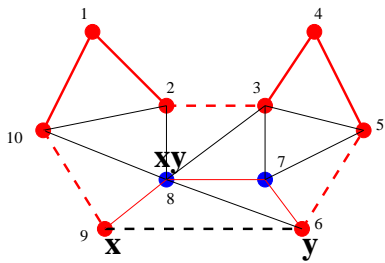}
\caption{An illustration of a triangulated graph and notations used. Edges in red delineate
the shape. Dashed edges are removable. Red vertices are on the boundary of
the current triangulated graph. Blue vertices are internal. For each boundary, there is
a unique triangle associated with it.}
\label{fig:GraphIllustration}
\end{figure}

See Figure \ref{fig:GraphIllustration} for illustration. A
Hamiltonian cycle of the shape is delineated in red. Starting
from the top-left vertex, the order of the cycle is enumerated.
There are 8 boundary edges, which are shown with thicker lines.
Among 10 vertices, 8 are currently at the boundary of the
triangulated surface. They are shown with red disks. There are
two internal vertices and they are shown with blue disks. Thus,
$\textbf{B}=[\textbf{b}_1,\cdots,\textbf{b}_8]$ is
$[\textbf{p}_1,\cdots,\textbf{p}_6,\textbf{p}_9,
\textbf{p}_{10}]$ or its circularly rotated version. Currently,
there are 4 removable edges. They are shown with dashed lines.
Consider the removable edge at the bottom of the figure and let
the two vertices \textbf{x} and \textbf{y}. Then,
$\textbf{xy}=\textbf{p}_8$. At this point,
$(\textbf{x},\textbf{y})$ has the largest flatness measure
among the four removable edges. After $(\textbf{x},\textbf{y})$
is removed, $(\textbf{p}_2,\textbf{p}_3)$ and
$(\textbf{p}_9,\textbf{p}_{10})$ are no longer removable.
Instead, we have two removable edges:
$(\textbf{p}_6,\textbf{p}_8)$ and
$(\textbf{p}_5,\textbf{p}_6)$. We will recover the Hamiltonian
cycle shown in red by removing $(\textbf{p}_6,\textbf{p}_8)$.
We will recover a different Hamiltonian cycle by removing
$(\textbf{p}_5,\textbf{p}_6)$.

\begin{figure}
\centering
\includegraphics[width=2.0in]{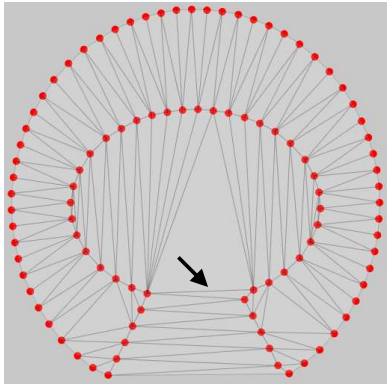}
\caption{An example shape that may be problematic for Algorithm 1. The edge pointed by the arrow, when it becomes a boundary edge
after triangles below it are removed, will have the flatness of 0. Since the circumscribing circle of the
associated triangle (the large one with the arrow in it) does not cross the shape, the triangle persists regardless of $K$.
Thus, the flatness of the boundary edge remains 0, which makes removing it based on the flatness measure problematic.}
\label{fig:Ushape}
\end{figure}

Note that the flatness measure captures the heuristic described
previously. If $\textbf{x}$ and $\textbf{y}$ are adjacent pair,
$\lim_{K\rightarrow\infty}\|\textbf{x}-\textbf{y}\|=0$. Thus,
$\lim_{K\rightarrow\infty}\phi(\textbf{x},\textbf{y})=0$. On
the other hand, when $\textbf{x}$ and $\textbf{y}$ are not
adjacent pair, but either $\textbf{x}$ and $\textbf{xy}$ or
$\textbf{y}$ and $\textbf{xy}$ is, then
$\lim_{K\rightarrow\infty}\|\textbf{x}-\textbf{y}\|>0$ and
either $\lim_{K\rightarrow\infty}\|\textbf{x}-\textbf{xy}\|=0$
or $\lim_{K\rightarrow\infty}\|\textbf{y}-\textbf{xy}\|=0$.
Thus, $\lim_{K\rightarrow\infty}\phi(\textbf{x},\textbf{y})>0$.
The remaining case is when none of $(\textbf{x},\textbf{y})$,
$(\textbf{x},\textbf{xy})$, and $(\textbf{y},\textbf{xy})$ are
adjacent. This case is rare but can happen around a concave
part of the shape. We claim that, in this case,
$\|\textbf{x}-\textbf{y}\|$ tends to be the largest among the
three sides of the triangle. Thus,
$\phi(\textbf{x},\textbf{y})>0$. This is because Delaunay
triangulation prohibits a point to be inside the circumscribing
circle of any triangle. When $\|\textbf{x}-\textbf{y}\|$ is the
largest, the center of the circumscribing circle will be
situated away from the shape. Then, it has good chance that no
parts of the shape are inside the circle. When
$\|\textbf{x}-\textbf{y}\|$ is not the largest, the center of
the circumscribing circle will be situated toward the shape.
Then, it has good chance that some parts of the shape are
inside the circle. As $K$ increases, a point from such part
will be sampled, and the triangle will be eliminated from the
triangulation. Of course, we can carefully construct a shape
where $\|\textbf{x}-\textbf{y}\|$ is the smallest, yet the
circumscribing circle never crosses the shape. For example, see
Figure \ref{fig:Ushape}. A large triangle with the arrow in it
persists as $K$ increases while the edge at the base of the
triangle has the flatness of 0. In such cases, we need a
tie-breaker that choose an edge with non-adjacent points over
those with adjacent ones, because both have the flatness of 0.
The length of the boundary edge is a possible tie-breaker as it
remains relatively unchanged as $K$ increases while a boundary
edge formed by adjacent points approaches zero. Using the
tie-breaker, we can indeed remove the edge and successfully
extract the complete shape of Figure \ref{fig:Ushape}. However,
since shapes like the one shown in Figure \ref{fig:Ushape} is
rare, we do not explore such option in our experiments and keep
the flatness measure to the simple form of (\ref{eq:flatness}).

Now, we describe an algorithm to remove triangles outside the
shape using the flatness measure of (\ref{eq:flatness}). We
want to do it in a determinant fashion so that we can easily
reproduce the results. We want to do it in a computationally
efficient and simple manner so that the performance is impacted
by the representation and not by sophisticated search and
optimization techniques. More specifically, we limit ourselves
to a polynomial algorithm without any backtracking. This led to
a greedy algorithm shown in Figure \ref{algo:Algorithm}.

Note that Delaunay triangulation of $K$ points takes $O(K\log
K)$. There are $O(K)$ edges in the triangulation, thus
collecting removable boundary edges takes $O(K)$ and sorting
them takes $O(K\log K)$. Since each edge is visited at most
once in the while loop and insertion of edges into the sorted
list takes $O(\log K)$, the loop takes $O(K\log K)$. Sequencing
the boundary points takes $O(K)$. Thus, the overall complexity
of the algorithm is $O(K\log K)$. Note that
$(\textbf{x},\textbf{y})\in\textbf{E}$ can change from
removable to unremovable since removing another edge can expose
$\textbf{xy}$. Also, $(\textbf{x},\textbf{xy})$ or
$(\textbf{y},\textbf{xy})$ may not be removable when they are
inserted in \textbf{E}. We could check the removable condition
before adding them to $\textbf{E}$ to speed up the computation
slightly. We did not do so in the algorithm to keep it simple,
as it does not affect the overall complexity.

\begin{algorithm}[h]
\KwIn{$\textbf{P}$: a (non-sequenced) set of sample points}
\KwOut{$\textbf{B}$: a sequenced series of boundary points}
Triangulate \textbf{P}\\
Collect removable boundary edges and store them in \textbf{E}.\\
Sort edges in \textbf{E} according to (\ref{eq:flatness}) in the descending order.\\
\While{$\textbf{E}\ne\emptyset$} {
      Pop $(\textbf{x},\textbf{y})$ from \textbf{E}\\
      \If{\textbf{xy} is internal}{
          Insert $(\textbf{x},\textbf{xy})$ and $(\textbf{y},\textbf{xy})$ to \textbf{E} according to
          (\ref{eq:flatness}).\\
      }
} Sequence the boundary points.\caption{Algorithm1:
Surface-based Grouping Algorithm}\label{algo:Algorithm}
\end{algorithm}

Next, we design a similar grouping algorithm for the contour
based model. As stated in Introduction, most contour based
grouping algorithms require tangential and/or curvature
information, thus are not applicable for our purpose. By the
same reasoning to keep our surface based algorithm simple, we
want a simple algorithm that is not heavily dependent on
optimization and elaborate searching. We decided to use a
Minimum Spanning Tree (MST) to group sampled points using the
proximity information. Basically, we form a graph in which
sampled points are vertices and every pair of vertices is
connected by an edge whose weight is set to the Euclidean
distance between the vertices. We then apply Kruskal's MST
algorithm to find an MST \cite{cormen2001introduction}. A
straightforward implementation of Kruskal with a fully
connected graph has the computational complexity of $O(K^2\log
K$).

One issue of an MST is that it can have branches, thus we may
not be able to translate the tree into a simple cycle easily.
Instead of comparing two methods based on the extracted shape,
we use edges selected by them to score the performance. The
original shape can be represented by $K$ edges that links
adjacent points. Each method yields a set of edges: at most $K$
edges for the surface based one and $K-1$ for the contour based
one. We use the proportion of edges that are in the original
shape as the performance measure. More precisely, the grouping
score denoted as $\xi$ is defined as
\begin{equation}\label{eq:GroupingScore}
\xi\left(\textbf{E}_K\right)=
\frac{|\textbf{E}_K\cap\textbf{E}(\textbf{P})|}{|\textbf{E}_K|}
\end{equation}
where $\textbf{E}_K$ is a set of boundary edges for the surface
based method and a set of MST edges for the contour based
method at the sample size of $K$, $\textbf{E}(\textbf{P})$ is a
set of edges formed by connecting every circularly adjacent
points in $\textbf{P}$, and $|\textbf{E}|$ gives the number of
edges in \textbf{E}. Note that $\textbf{E}(\textbf{P})$ has $K$
edges, delineates the original shape, and serves as the ground
truth in the measure.

Note that the contour based method benefits more by this
measure. The number of edges in an MST is $K-1$ instead of $K$,
thus a result with the perfect score of 1 still does not
extract the entire shape. More importantly, the MST often has
branches, which results in a perceptually unacceptable
interpretation of the shape. Yet, it may result in a high
score. For the surface based one, the solution is always a
simple cycle, and the quality of the solution and the
performance measure typically match up well.

The second part of the study is to investigate how the grouping
performance of the algorithms compare to the recognition
performance presented in Experiment 1. To do so, we need to
build a shape retrieval algorithm on top of the grouping
result. There have been many algorithms to retrieve a shape
from a database using contour features. Earlier ones only allow
rigid transformation on the contour
\cite{Asada:PAMI86,Mokhtarian:Shape}. For an extensive review,
see \cite{zhang2004review}. More recent ones allow non-rigid
transformation and partial occlusion
\cite{Sato:IJCV1998,Sebastian:PAMI2003,Mio:IJCV2007}. Since our
goal is not to develop a full-fledged shape retrieval system,
we keep the shape retrieval algorithm simple. We decided to use
DC-normalized Fourier descriptors computed on centroid
distances as the contour features \cite{Gonzalez:DIP} and use
the Euclidean distance in the feature space for a similarity
measure between features.

The DC-normalized Fourier descriptors are obtained from a
sequence of points delineating the shape by first computing the
distance from the centroid of the sequence of points to each
point, computing the Fourier coefficients of the real valued
distance values, taking the absolute value of each complex
valued Fourier coefficient, and then dividing each absolute
coefficient by the DC component. We divide each coefficient by
the DC component so that the descriptors are scale invariant.
We use the centroid distance as it is shown more effective than
the raw coordinate \cite{zhang2001comparative}.

Using the DC-normalized Fourier descriptors, we compute the
sample size needed to retrieve each reference shape from the
collection of 20 shapes. Figure \ref{algo:RetrievalAlgorithm}
summarizes the retrieval algorithm. It takes the database of
shapes ($Db$) and an index ($id$) to a reference shape in the
database. We pre-compute the DC-Normalized Fourier descriptors
for each shape $\textbf{Z}_i$ and store them in the database.
We only store 10 coefficients corresponding to the 10 lowest
frequency components, as they tend to capture the principle
structures of the shape while higher frequency components tend
to capture perturbation, noise, and artifacts.

The algorithm loops over $K$ from 30 at increment of 10, like
in Experiment 1 (Line 1). Since we need at least 21 sample
points to collect 10 DC normalized Fourier coefficients, we
start from 30 instead of 10 as in Experiment 1. For each $K$,
we sample the shape (Line2), apply Algorithm1 to extract a
sequence of boundary points (Line 3), and compute 10 Fourier
descriptors (Line 4) using the boundary points. We compare the
descriptors with those in the database using a simple Euclidean
distance (Lines 5-6). Note that $|Db|$ in Line 5 gives the
cardinality of the database (20 in our case). We deemed the
shape is \textit{retrievable} if the Euclidian distance of the
correct shape ($s_{id}$) is the smallest and the next smallest
is at least 3 times larger. (Line 7). As soon as the shape
becomes retrievable, the algorithm returns the sample size
denoted as $n$ (Line 8). Although the algorithm always
terminated in our experiments, the termination condition may
never be met if two shapes in the database are very similar. In
this simple implementation, we do not worry about such
situation. The algorithm mimics the behavior of a careful human
observer who commits to an answer only when he/she is very
certain about the choice (three times more than the other
shapes).

\begin{algorithm}[h]
\KwIn{$Db=\left\{\left(\textbf{S}_i,\textbf{Z}_i\right)\right\}$:
the shape database with both shapes and Fourier descriptors.}
\KwIn{$id$: a reference shape id}
\KwOut{$n$: Retrievable sample size}
\ForEach{$n=30,40,50,\cdots$}{
    \textbf{P}=Sample $n$ points from $\textbf{S}_{id}$\\
    \textbf{B}=Algorithm1(\textbf{P})\\
    $\textbf{x} = FourierDescriptorsOf(\textbf{B},10)$\\
    \For{i=1 \emph{\KwTo} $|Db|$}{
        $s_i=\|\textbf{x}-\textbf{Z}_i\|$\\
    }
    \If{$s_{id}=\min_i(s_i)$ {\bf and} $\min_{i\ne
    id}\left(s_i/s_{id}\right)>3$}{\Return $n$}
} \caption{Algorithm2: Shape retrieval algorithm}
\label{algo:RetrievalAlgorithm}
\end{algorithm}

Unfortunately, the above algorithm is not applicable to the
contour based grouping, which gives a tree not a shape. In
order to derive some meaningful quantities associated with the
contour based approach, we devise another way to relate the
grouping performance to a retrieval one. For each shape, we
examine grouping results at various $K$. We start from $K=10$
and increase $K$ at 10 increment until $K=200$. For each $K$,
we evaluate the grouping score, $\xi$, defined previously.
Then, we find a point in $K$ denoted as $m$ after which the
grouping score remains above 80\%. The underlying assumption is
that shape retrieval becomes trivial when the grouping accuracy
reaches a certain level. We chose 80\% empirically but observed
that the overall result was not sensitive to the choice. With
this measure, we can compare the contour based algorithm with
the human performance as well as the surface based algorithm.

\section{Results}\label{results}
\subsection{Experiment 1}
Among six participants in the $\Omega_T$ group, two did not use
SPACE bar at all and guessed the shape from the initial
representation at $K=10$. Their guesses were almost all
incorrect, and therefore, we excluded their results in our
analysis. Other participants answered most shapes correctly. We
did not remove occasional incorrect answers from our results,
because answering 'fish' instead of 'dolphin' or 'plane'
instead of 'duck' shows some degree of recognition. Thus, there
are 8 participants in $\Omega_P$ group, 6 in $\Omega_A$ group,
and 4 in $\Omega_T$ group, and each participant has 20 data
points.

The mean and the standard deviation of $K$ for each group and
shape are listed below in Table \ref{table:Result1}. In
general, $\Omega_T$ has the smallest mean, followed by
$\Omega_P$, and $\Omega_A$ having the largest mean. According
to the Kruskal-Wallis test, the representation of shapes
affected the response time significantly ($H(2)=43.16$,
$p<0.01$).

\begin{figure}
\centering
\includegraphics[width=3.0in]{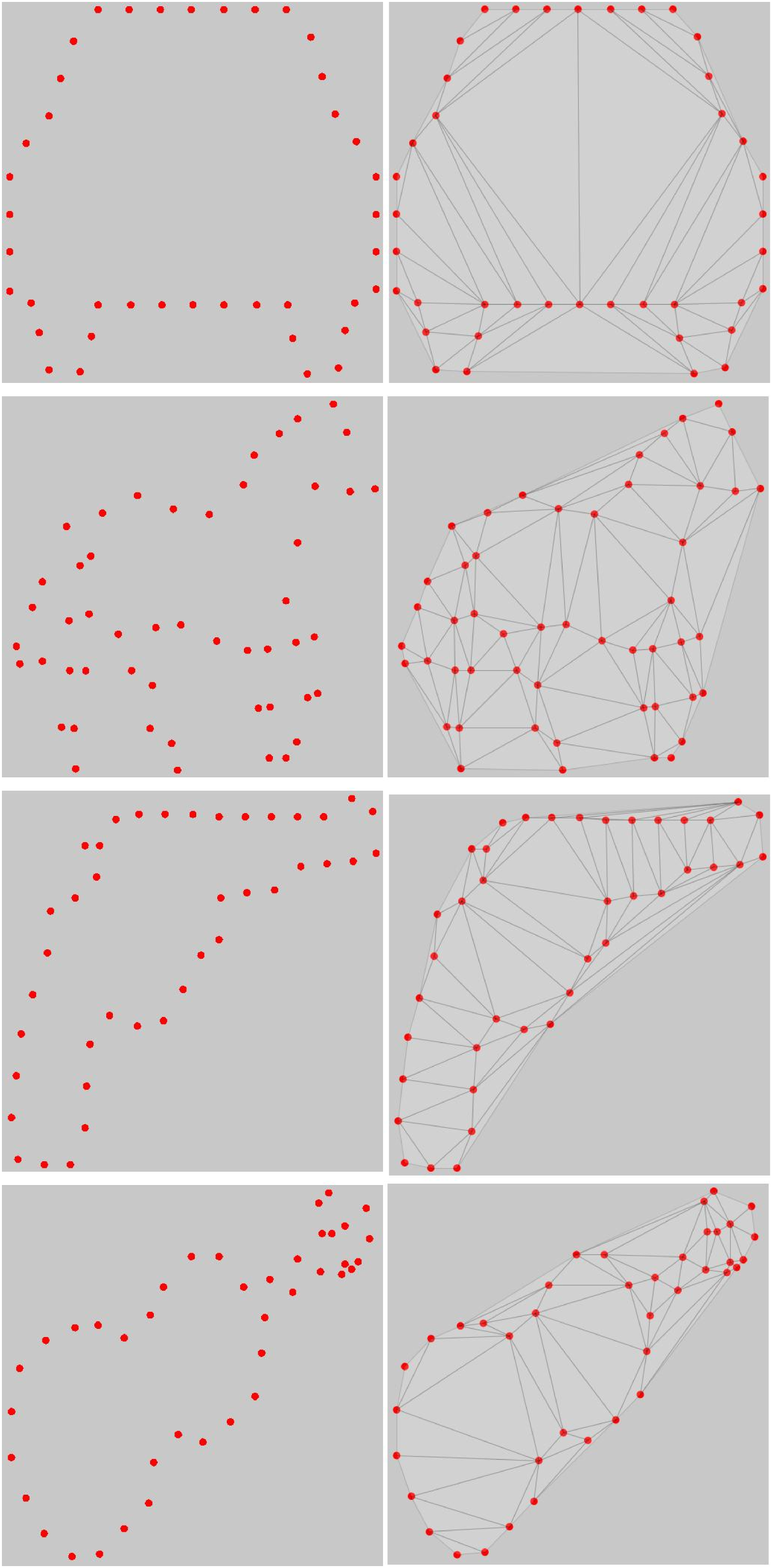}
\caption{Four shapes (Car, Horse, Pistol, and Violin)
whose results were significantly different between
$\Omega_P$ (Left) and $\Omega_A$ (Right).}
\label{fig:AllTriangleDiscrepancy}
\end{figure}

Next, we investigate more closely how recognition of each shape
is affected by different representations. Table
\ref{table:Result1_t} shows p-values of pooled t-test between
$\Omega_P$ and $\Omega_A$, and $\Omega_P$ and $\Omega_T$,
treating $\Omega_P$ as the control group. Those entries with
$p<0.05$ are shown in bold. There are four shapes in $\Omega_P$
vs. $\Omega_A$ column that are $p<0.05$. Figure
\ref{fig:AllTriangleDiscrepancy} shows these four shapes in
$\Omega_P$ and $\Omega_A$ at $K$ near the mean value for
$\Omega_P$. Between $\Omega_P$ and $\Omega_T$, seven shapes
show significant difference. For six shapes out of the seven,
$\Omega_T$ shows faster recognition than $\Omega_P$. An
exception is the umbrella shape where the triangulation failed
to represent accurately the handle of the umbrella, making the
recognition of the shape difficult. In general, grouping
provided by $\Omega_T$ facilitated recognition. The results
also indicate that the triangulation retained the correct
grouping of cues in most cases.

The results show that the representations affect recognition,
which contradicts our initial hypothesis that our visual system
groups a set of dots via triangulation. However, the results do
not reject a more general surface based grouping hypothesis.
Furthermore, the results suggest that our visual system does
not depend exclusively on either points or triangles. If our
recognition only uses points representation, adding triangles
as in $\Omega_A$ would not have had any effect on recognition.
If our recognition only uses triangles, the recognition time of
all three representations would have been similar and our
initial hypothesis would have not been rejected. Thus, the
results only paint a more complex picture of our visual system
that combines various cues to facilitate the processing, even
at a very basic level of grouping of dots; some configurations
of triangles facilitate and some distract the grouping process.

\begin{figure}
\centering
\includegraphics[width=2.0in]{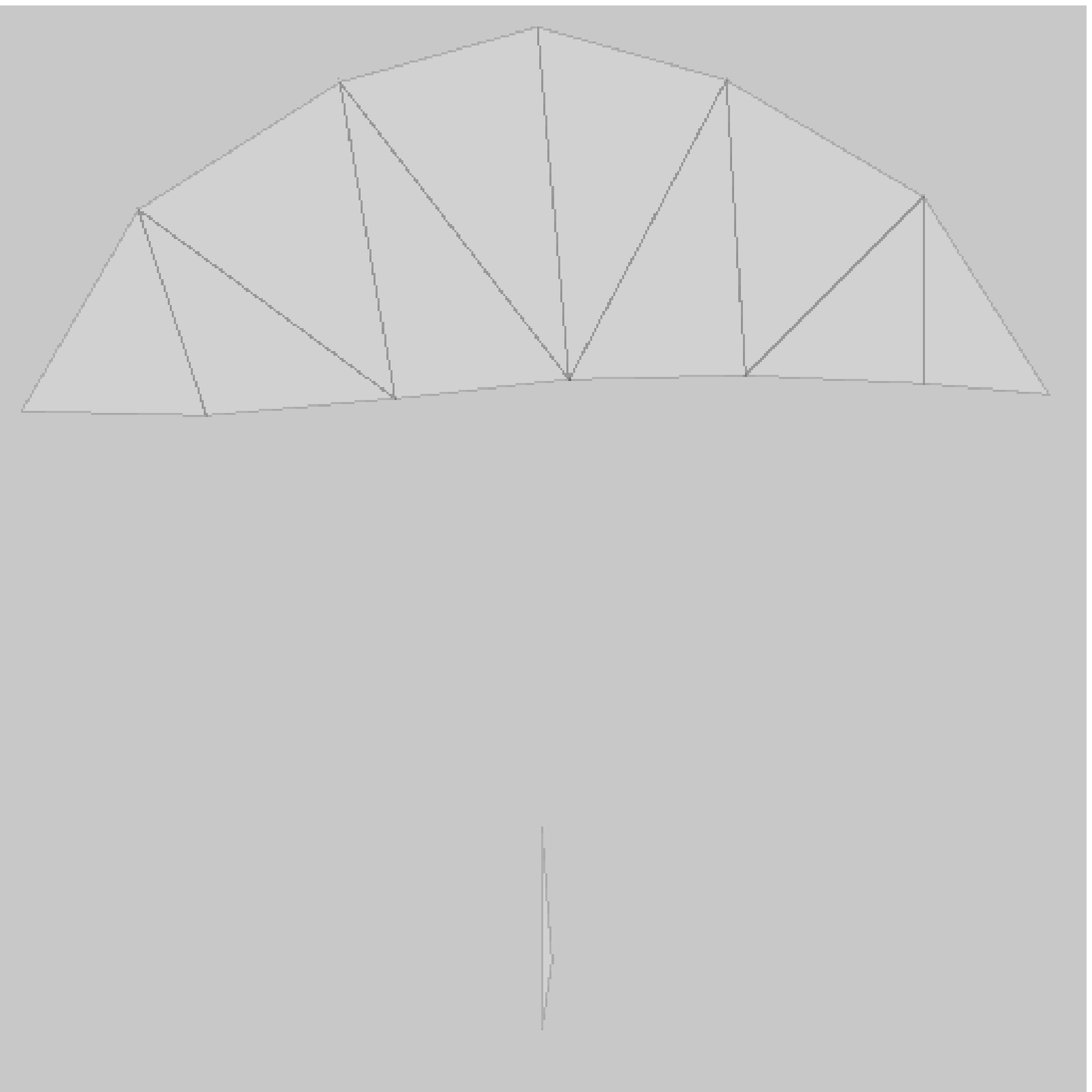}
\caption{An example of parts based shape recognition. At $K=20$, parts are clearly
disjoint. However, some participants were able to recognize the shape (Umbrella) correctly.}
\label{fig:umbrella20}
\end{figure}

An interesting observation we were able to make was that
recognition was often parts based. This is clear from the
umbrella result of $\Omega_T$. Some participants were able to
recognize the umbrella correctly at $K=20$. See Figure
\ref{fig:umbrella20} for the image at $K=20$. At this point,
only a small part of the pole/handle was visible and was
disjoint with the canopy. Thus, in some cases, recognition came
before reconstruction of the shape.

The part based recognition may be a reason for slow recognition
of shapes shown in Figure \ref{fig:AllTriangleDiscrepancy}. In
these shapes, some important parts were obscured by triangles.
For the Car shape, we think that wheels are important parts. In
$\Omega_A$, they were obscured by clusters of similar triangles
surrounding them. For the Horse shape, we think that legs are
important parts. In $\Omega_A$, they were distracted by
accidental alignment of triangles across legs. For the violin
shape, we think that the body is an important part. In
$\Omega_A$, it was obscured by triangles encompassing smoothly
over the entire instrument. For the pistol, we think that the
trigger and the hammer are important parts. In $\Omega_A$, they
were obscured by clusters of nearby triangles.

\begin{table}
\centering \caption{Mean sample size at recognition of shapes
($K$) for each representation and each shape.}
\begin{tabular}{|c|c|c|c|c|c|c|}
  \hline
  &\multicolumn{2}{|c|}{$\Omega_P$} & \multicolumn{2}{|c|}{$\Omega_A$} &\multicolumn{2}{|c|}{$\Omega_T$}\\
  \hline
    Shape & Mean & Std & Mean & Std & Mean & Std \\
  \hline
Anchor & 32.50 & 6.61 & 36.67 & 14.91 & 22.50 & 5.00 \\
Arrow & 31.25 & 11.66 & 33.33 & 9.43 & 20.00 & 0.00 \\
Balloon & 37.50 & 15.61 & 35.00 & 12.58 & 25.00 & 10.00 \\
Camel & 43.75 & 15.76 & 65.00 & 23.63 & 42.50 & 5.00 \\
Car & 40.00 & 5.00 & 63.33 & 11.06 & 52.50 & 20.62 \\
Cherries & 41.25 & 9.27 & 51.67 & 13.44 & 45.00 & 17.32 \\
Dolphin & 40.00 & 8.66 & 31.67 & 3.73 & 20.00 & 0.00 \\
Duck & 40.00 & 10.00 & 48.33 & 12.13 & 35.00 & 17.32 \\
Fish & 45.00 & 18.71 & 43.33 & 7.45 & 22.50 & 5.00 \\
Horse & 45.00 & 8.66 & 61.67 & 14.62 & 37.50 & 5.00 \\
Kangaroo & 58.75 & 9.27 & 56.67 & 13.74 & 35.00 & 17.32 \\
Palm tree & 61.25 & 7.81 & 56.67 & 7.45 & 37.50 & 5.00 \\
Pistol & 42.50 & 7.07 & 86.67 & 24.27 & 45.00 & 19.15 \\
Plane & 41.25 & 8.35 & 46.67 & 9.43 & 35.00 & 12.91 \\
Shoe & 33.75 & 4.84 & 36.67 & 7.45 & 22.50 & 5.00 \\
Squirrel & 46.25 & 9.92 & 50.00 & 5.77 & 32.50 & 9.57 \\
Star & 23.75 & 6.96 & 30.00 & 11.55 & 15.00 & 5.77 \\
Telephone & 27.50 & 6.61 & 36.67 & 12.47 & 17.50 & 5.00 \\
Umbrella & 18.75 & 3.31 & 38.33 & 28.53 & 30.00 & 11.55 \\
Violin & 38.75 & 3.31 & 46.67 & 4.71 & 40.00 & 14.14 \\
  \hline
All & 39.50 & 13.91 & 47.75 & 19.77 & 31.62 & 14.36 \\
  \hline
\end{tabular}\label{table:Result1}
\end{table}

\begin{table}
\centering \caption{p-values of Pooled t-tests for each shape
in Experiment 1. Those less than 0.05 are shown in bold.}
\begin{tabular}{|c|c|c|}
  \hline
    Shape & $\Omega_P$ vs $\Omega_A$ & $\Omega_P$ vs $\Omega_T$\\
  \hline
Anchor & 0.53 & \textbf{0.03} \\
Arrow & 0.75 & 0.11 \\
Balloon & 0.77 & 0.20 \\
Camel & 0.09 & 0.89 \\
Car & \textbf{4e-4} & 0.12 \\
Cherries & 0.14 & 0.64 \\
Dolphin & 0.06 & \textbf{2e-3} \\
Duck & 0.22 & 0.55 \\
Fish & 0.85 & 0.06 \\
Horse & \textbf{0.03} & 0.17 \\
Kangaroo & 0.76 & \textbf{0.01} \\
Palm tree & 0.33 & \textbf{4e-4} \\
Pistol & \textbf{6e-4} & 0.93 \\
Plane & 0.30 & 0.33 \\
Shoe & 0.43 & \textbf{5e-3} \\
Squirrel & 0.46 & 0.05 \\
Star & 0.27 & 0.07 \\
Telephone & 0.13 & \textbf{0.03} \\
Umbrella & 0.10 & \textbf{0.03} \\
Violin & \textbf{0.01} & 0.81 \\
  \hline
\end{tabular}\label{table:Result1_t}
\end{table}

\section{Experiment 2}
\begin{figure}
\centering
\includegraphics[width=4.0in]{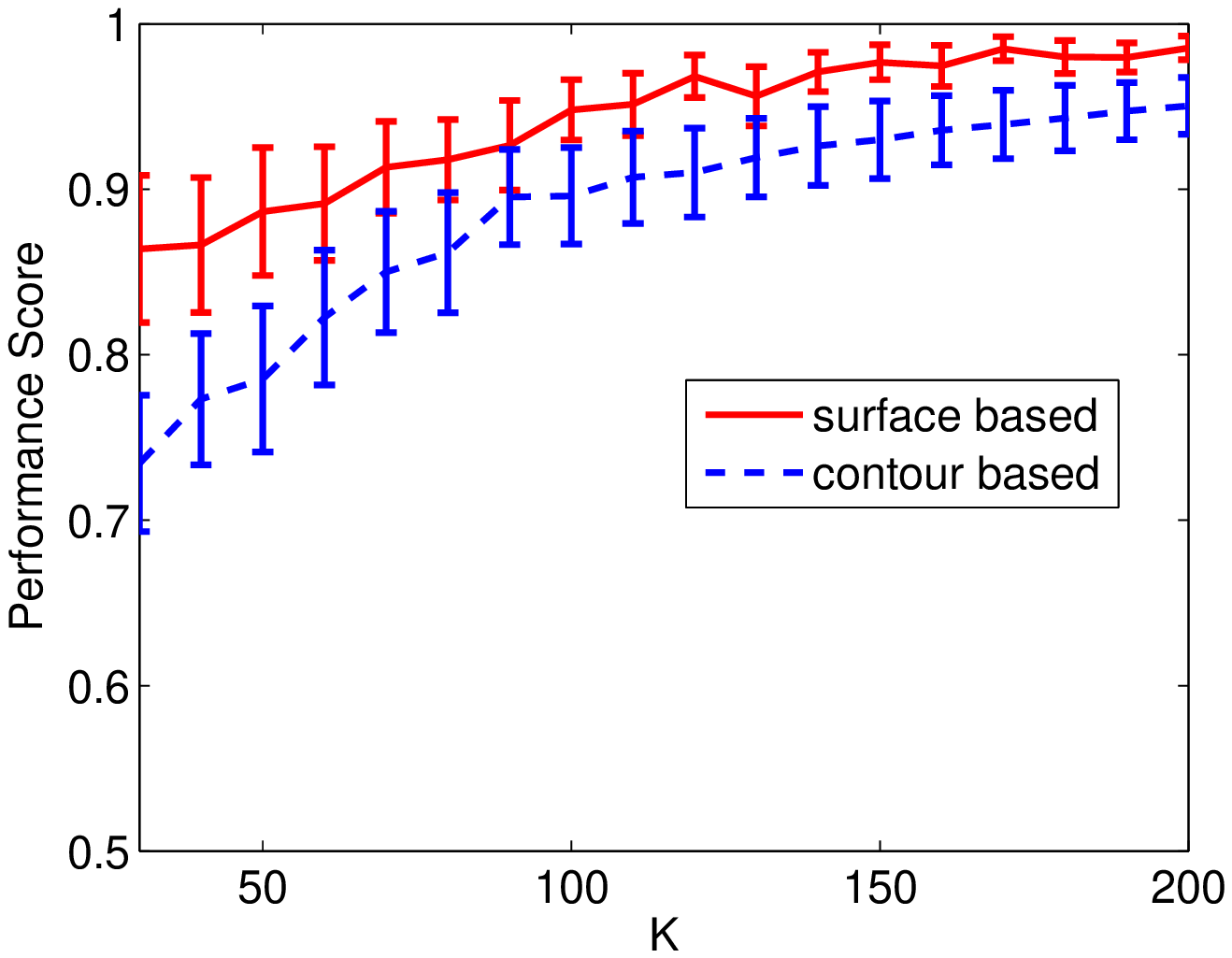}
\caption{Performance comparison between the surface based and contour based grouping algorithms.
Mean values of $\xi$ over 20 shapes are plotted at different $K$. The error bars are standard error of
the mean.}
\label{fig:PerformanceScorePlots}
\end{figure}

First, we compare the performance score $\xi$ defined in
(\ref{eq:GroupingScore}) between the surface based algorithm
and the contour based one at different $K$. Figure
\ref{fig:PerformanceScorePlots} shows plots of the mean score
over 20 shapes at $K=30, 40, \cdots, 200$. The accompanying
error bars are the standard error of the mean. Clearly, the
surface based method consistently provides more accurate
grouping of cues than the contour based one.

\begin{figure}
\centering
\includegraphics[width=4.0in]{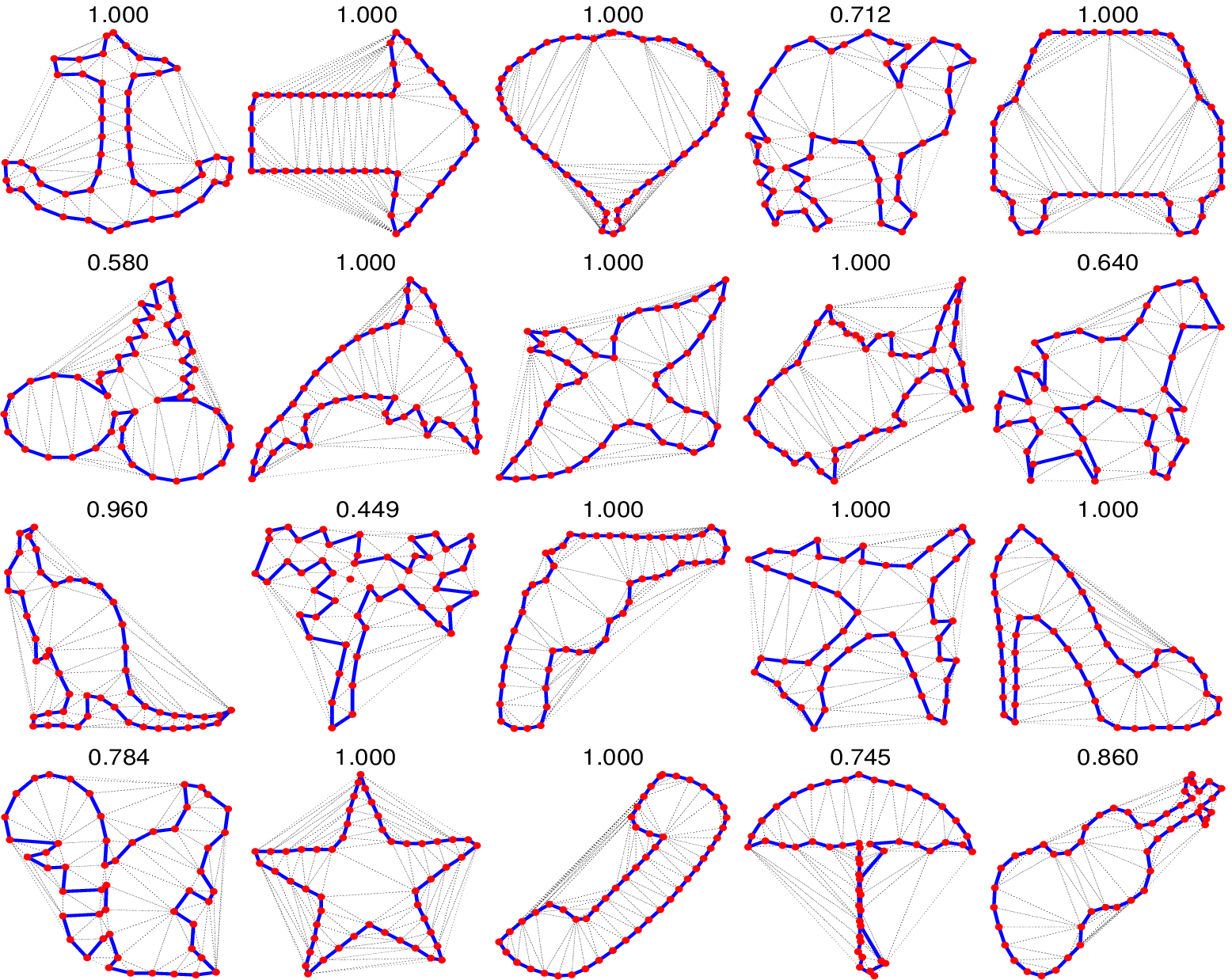}
\caption{Results of the surface based grouping algorithm at $K=50$. Edges selected
are shown in blue. Dashed gray edges show triangulation. The number at the top of each result is the
value of $\xi$.}
\label{fig:surfaceResults}
\end{figure}

\begin{figure}
\centering
\includegraphics[width=4.0in]{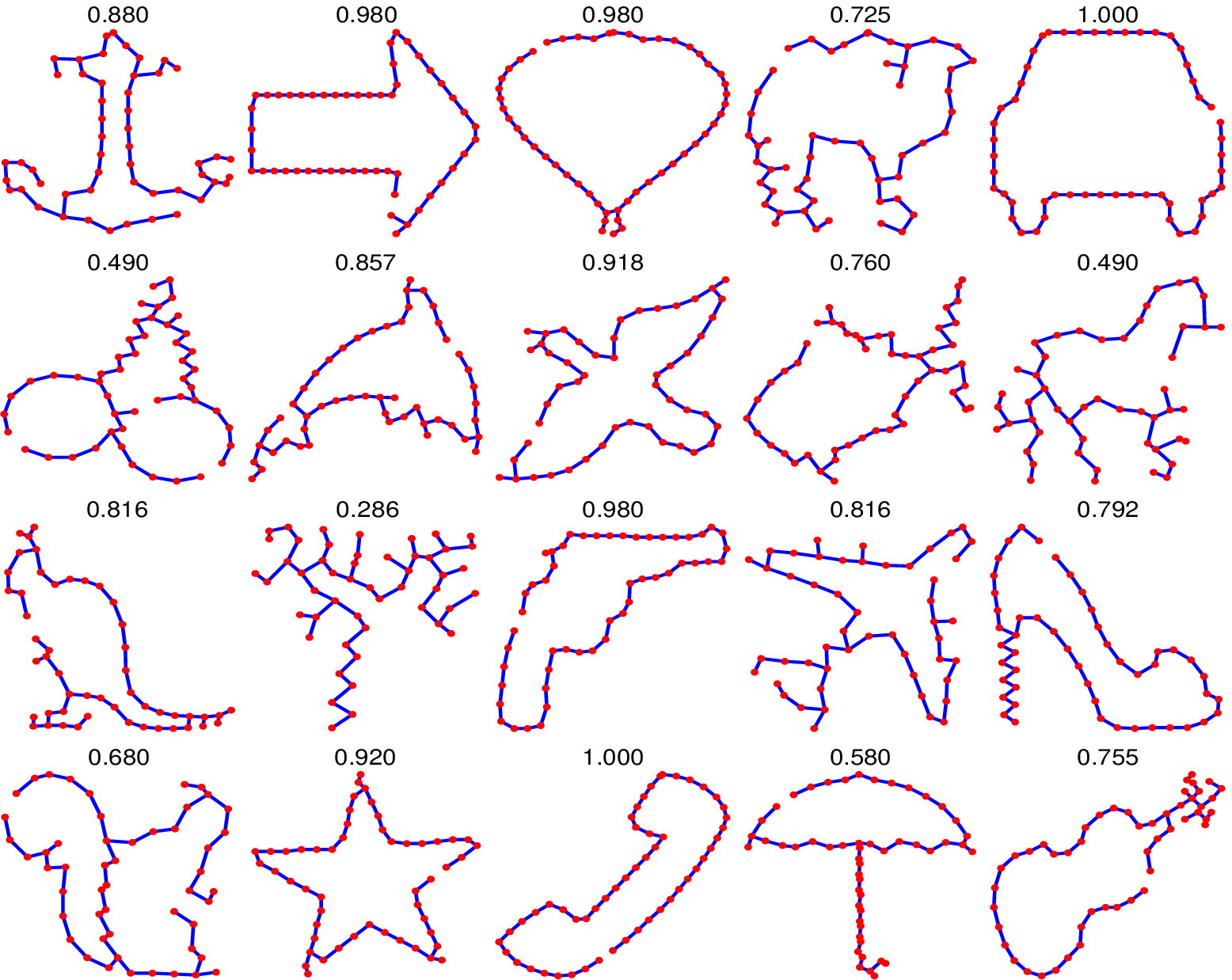}
\caption{Results of the contour based grouping algorithm at $K=50$. Edges selected
are shown in blue. The number at the top of each result is the
value of $\xi$.}
\label{fig:contourResults}
\end{figure}

Figures \ref{fig:surfaceResults} and \ref{fig:contourResults}
show instances of shape extraction at $K=50$ with the surface
based approach and the contour based approach, respectively.
Results of 20 shapes are shown. In each figure, sample points
are shown with red dots, and the grouping result is shown with
blue lines. In Figure \ref{fig:surfaceResults}, the
triangulation of the sample points is also shown with dashed
lines. The number above each figure is the value of $\xi$. The
surface based method was able to find a Hamiltonian cycle on 19
shapes out of 20. One exception was the palm tree shape. As
stated in Method Section, the contour based method, which finds
a minimum spanning tree, often finds a tree with branches,
making translation of the tree into a simple cycle difficult.

The surface based method showed difficulty dealing with a shape
with thin structures. This is largely due to Delaunay
triangulation's preference toward fat triangles over thin ones.
This characteristic is evident in the handle of Umbrella shape
and the stems of Cherry shape. The Cherry shape actually
contain two objects (two cherries), which poses another
problem; Algorithim1 at the current form is not capable of
dealing with multiple objects.

Table \ref{table:Result2} shows the retrievable sample size $n$
and $m$ for each shape and for each grouping method (surface
based and contour based ones). Note that $n$ is the result of
Algorithm 2 shown in Figure \ref{algo:RetrievalAlgorithm}, and
$m$ is the sample size that guarantees 80\% grouping accuracy
as described in the Method Section. Algorithm 2 is not
applicable to the contour based results.

Table \ref{table:Result2b} shows Pearson correlation
coefficients and their p-values between an algorithmic
performance measure ($n$ and $m$) and a mean recognition speed
as reported in Table \ref{table:Result1} with various
representation types ($\Omega_P$, $\Omega_A$, and $\Omega_T$).
Note that $n$ of the surface based method has highest
correlation among the three algorithmic ones with all types of
the human retrieval performance.


\begin{table}
\centering \caption{The results of the shape retrieval studies.
$n$ is the result of the surface based retrieval algorithm
(Figure \ref{algo:RetrievalAlgorithm}), which is not applicable
to the contour based approach. $m$ is the minimum sample size
that guarantees $\xi>0.8$.}
\begin{tabular}{|c|c|c|c|c|}
  \hline
  &\multicolumn{2}{|c|}{Surface} & \multicolumn{2}{|c|}{Contour}\\
  \hline
    & n & m & n & m \\
  \hline
Anchor & 30 & 30 & - & 50 \\
Arrow & 30 & 30 & - & 30 \\
Balloon & 30 & 30 & - & 30 \\
Camel & 110 & 70 & - & 80 \\
Car & 30 & 30 & - & 30 \\
Cherries & 150 & 120 & - & 90 \\
Dolphin & 30 & 30 & - & 40 \\
Duck & 50 & 30 & - & 30 \\
Fish & 30 & 30 & - & 70 \\
Horse & 120 & 130 & - & 180 \\
Kangaroo & 40 & 30 & - & 50 \\
Palm tree & 150 & 100 & - & 140 \\
Pistol & 30 & 30 & - & 30 \\
Plane & 30 & 30 & - & 50 \\
Shoe & 30 & 30 & - & 60 \\
Squirrel & 120 & 60 & - & 70 \\
Star & 30 & 30 & - & 40 \\
Telephone & 30 & 30 & - & 30 \\
Umbrella & 70 & 60 & - & 200 \\
Violin & 40 & 30 & - & 60 \\
\hline
Mean & 59.00 & 48.00 & - & 68.00 \\  \hline
\end{tabular}\label{table:Result2}
\end{table}

\begin{table}
\centering \caption{Correlation between algorithmic retrieval
performance and mean recognition speed of humans. Shown are
Pearson correlation coefficient with its p-value inside
parentheses. The correlation is taken between the mean column
($\Omega_P$, $\Omega_A$, and $\Omega_T$) in Table
\ref{table:Result1} and each column (surface $n$, surface $m$,
and contour $m$) in Table \ref{table:Result2}.}
\begin{tabular}{|c|c|c|c|c|}
  \hline
  &\multicolumn{2}{|c|}{Surface} & \multicolumn{2}{|c|}{Contour}\\
  \hline
    & n & m & n & m \\
  \hline
$\Omega_P$ & 0.42 (0.06) & 0.31 (0.18) & - & 0.04 (0.86)\\
$\Omega_A$ & 0.34 (0.14) & 0.16 (0.32) & - & 0.13 (0.59)\\
$\Omega_T$ & 0.44 (0.05) & 0.40 (0.08) & - & 0.19 (0.40) \\
  \hline
\end{tabular}\label{table:Result2b}
\end{table}

\section{Discussion}
The results of Experiment 1 showed that triangulation can
sometimes distract recognition. Thus, the results did not
support our original conjecture. However, it is still possible
that we employ triangulation as an internal representation at a
subconscious level. Then, the representation would not distract
our recognition. Also, the results only concern triangulation
based representation, and do not reject a more basic hypothesis
of surface based interpolation. Maybe, triangulation is too
"artificial" and our perception employs a more flexible and
smoother interpolation scheme. The surface based representation
does offer various computational advantages over the contour
based one as demonstrated in Experiment 2.

Removal of outer triangles ($\Omega_T$) enhances recognition
while inclusion of outer triangles ($\Omega_A$) hinders
recognition. By removing outer triangles, $\Omega_T$ provides
explicit grouping information that is not available in
$\Omega_P$ nor $\Omega_A$. $\Omega_A$ does contain all the
information that is present in $\Omega_P$. It also contains
some 'noise' in the form of outer triangles. Our experiment
showed that the noisy information sometimes distracts viewers.

The results of Experiment 2 support utility of the surface
based grouping. A simple greedy algorithm proposed in this
paper successfully reconstructed 12 shapes perfectly out of 20
with 50 sample points (See Figure \ref{fig:surfaceResults}). On
the other hand, a similar contour based approach could recover
only two shapes (Figure \ref{fig:contourResults}). The superior
performance of the surface based approach is the result of
restricting a solution to a simple cycle one. The approach
effectively recovered a Hamiltonian cycle in 19 shapes out of
20. Notable disadvantages of the surface based approach at the
current form are that it has difficulty in extracting thin
parts, and is not applicable to separate multiple objects.

Delaunay triangulation constructs a planar graph and reduces
the number of edges in the fully connected graph from $O(K^2)$
to $O(K)$. The reduction of edges facilitated grouping, yet did
not degrade the performance. Thus, Delaunay triangulation
effectively retained the underlying shapes in our experiments.
This is true for 'fat' objects but may not be so for 'thin'
objects.

\begin{figure}
\centering
\includegraphics[width=5.0in]{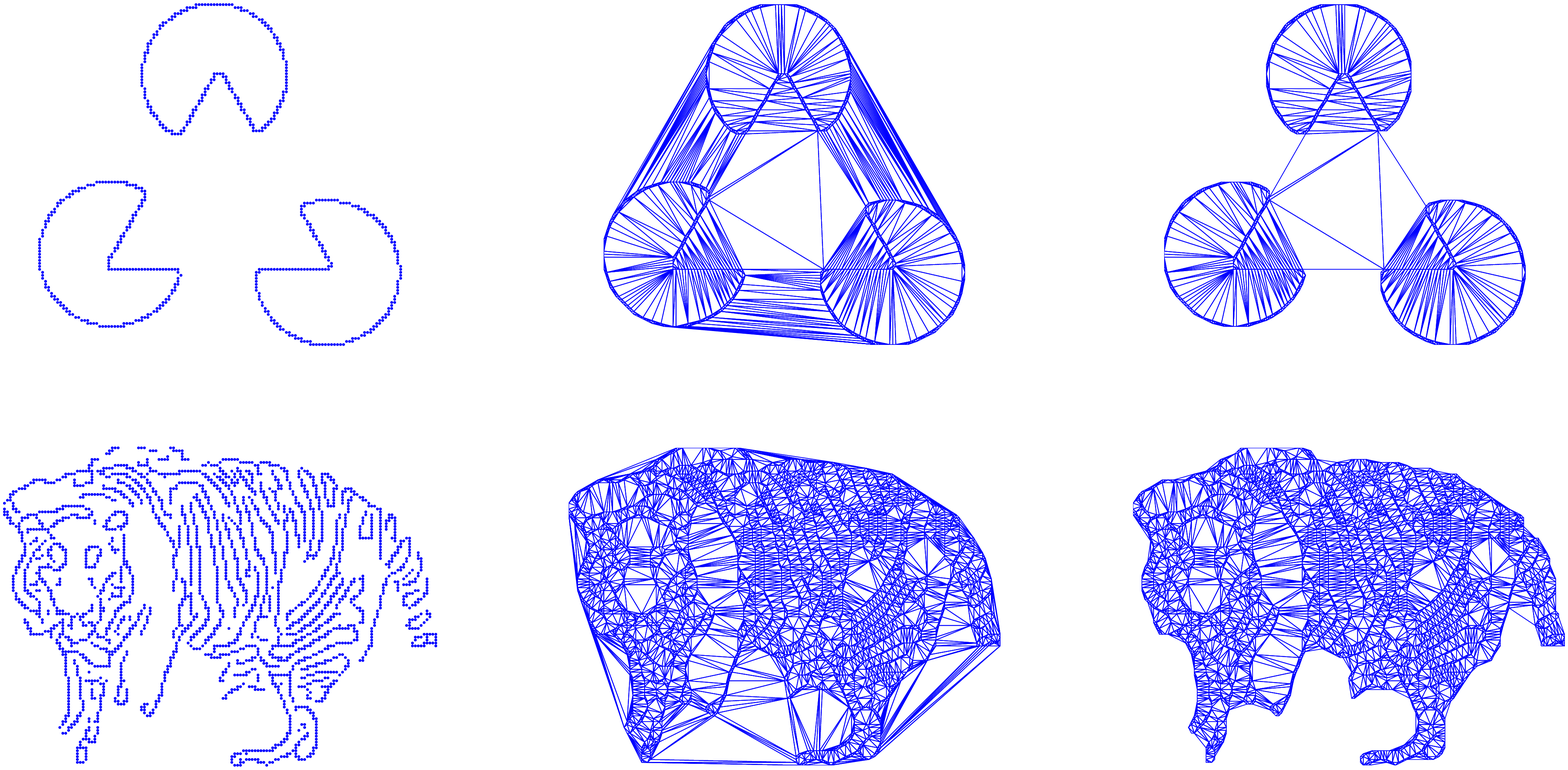}
\caption{Application of the algorithm with a stopping condition of $\phi>5$. Left: Edge images
obtained by Canny edge detector. Middle: Delaunay triangulation applied to the edge image. Right:
Remaining triangles after boundary edges with $\phi>5$ are removed incrementally.}
\label{fig:amodalExamples}
\end{figure}

Another advantage of the surface based approach is that it
tends to delineate illusory contours. Consider a problem where
we are given a set of edge points resulted from some edge
detector \cite{canny:edge,marr:vision} applied to an image with
an object. Our goal is to group these points into a coherent
shape outlining the boundary of the object. Since some edge
points represent patterns inside the object, the solution may
not be a Hamiltonian. Thus, instead of forcefully searching for
a Hamiltonian solution as in Figure \ref{algo:Algorithm}, we
stop the search when there is no boundary edges that are
'sufficiently' flat. We set the sufficiency to be 5 for our
illustration. Results of applying this modified algorithm to an
edge image of a Kanizsa triangle and another edge image with a
tiger are shown in Figure \ref{fig:amodalExamples}.

As the results show, the modified algorithm was able to produce
an illusory Kanizsa triangle. It was also able to group
disjoint groups of edge points at the tail of the tiger and
interpolate end points of occluded strip patterns on the body
of the tiger. These grouping tasks are rather difficult
computationally and various attempts have been made in the
past\cite{Saund:PO,geiger:ill,Sarti:Illusion} with highly
elaborate schemes. The surface based approach allows such
grouping to take place with a simple greedy approach. The
results also show some limitations of triangulation based
approach. In the Kanizsa triangle example, the pack-man shapes
are abruptly cut instead of being interpolated smoothly to
complete a disk. The triangle inside the illusory triangle is
artificially produced and does not agree with our perception.
We observed these issues in Experiment 1 where artificial lines
resulted from triangulation can potentially distract
recognition.

\begin{figure}
\centering
\includegraphics[height=1.5in]{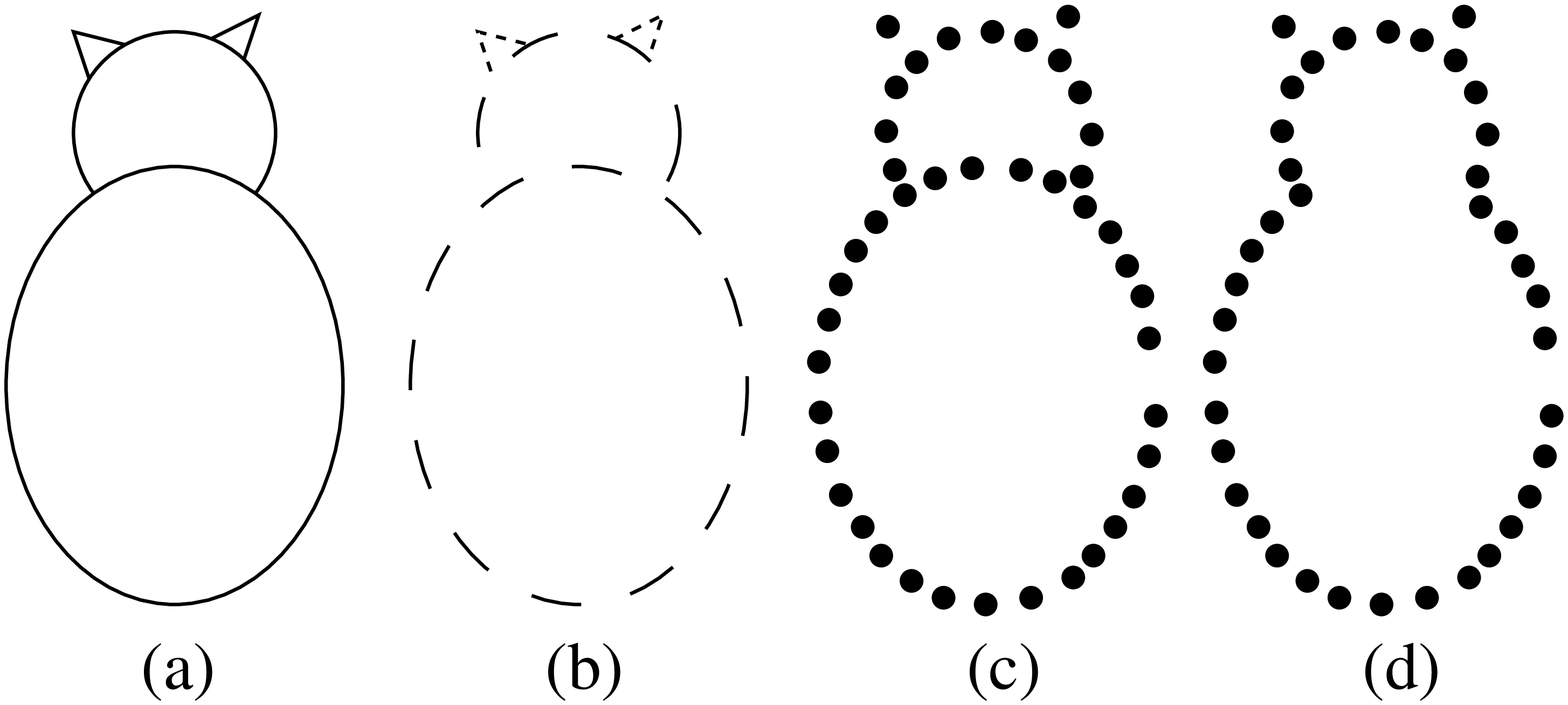}
\caption{Why do we see a bulge? (a) A cartoon drawing of a cat. We tend to see
a bulging 3D shape in both torso and head of this drawing. (b) The 3D perception tends to persist when gaps
are introduced to the boundary, (c) edge fragments are replaced with small disks, or (d)
the occlusion line is removed.}
\label{fig:Cats}
\end{figure}

The surface based approach also makes sense from evolutional
and ecological viewpoints, as the role of our vision is to
infer the 3D structure of the environment. Using surfaces to
construct the environment is more direct than using contours
first and then interpolate. We can also point out via a simple
drawing of Figure \ref{fig:Cats}(a) that we interpret this
drawing as a 3D figure rather than 2D drawing. In particular,
we tend to see a bulge in both torso and head of the cat like
figure. This perception persists when we introduce gaps as in
Figure \ref{fig:Cats}(b) replace contour fragments with
isolated dots as in Figure \ref{fig:Cats}(c), or further remove
the occlusion boundary between the head and torso as in Figure
\ref{fig:Cats}(d).

One explanation is that we recognize a cat from the drawing and
use our high-level knowledge of it to interpolate the surface.
However, no living cat actually looks like this drawing. One
may then argue that the drawing does not represent the actual
cat but a conceptual model of it. However, unless there has
been a link between this drawing to the actual cat, the
conceptual model cannot be developed\cite{Gibson:1951}. A more
plausible explanation in our opinion is that this drawing
although extremely simple (two ellipses and two triangles) has
enough information for us to infer most plausible surface
interpretation (perhaps in a Bayesian sense), which resulted
both ellipses to have a bulge. Using the interpretation, we
link it to a cat among all other things in our database, due to
features present in the drawing including the bulge. Once this
link is established, some features are further enhanced via the
conceptual model derived from the high-level knowledge.

Artists have developed numerous techniques to induce 3D
percepts in 2D drawings \cite{Gombrich:ArtAndIllusion}. Their
techniques demonstrate how easy it is to bring realistic
physical constructs from a drawing. Figure \ref{fig:Cats}(a) is
one simple example. Gibbon's ecological view suggests that a 3D
view of a form is primary while a 2D view is acquired via
training \cite{Gibson:1951}. Koenderink's catalog of surface
interpretations from contours and their joints provide some
clues as to why some features induce 3D percepts
\cite{Koenderink:1979}. Tse proposed volume based grouping of
contour fragments to account for amodal completion and extended
the relatability proposal of Kellman and
Shipley\cite{kellman:psycho,Tse:1999volume}. However, there is
no computational algorithm that groups a set of cues and
renders 3D interpretation possible. Since our triangulation
based approach did not agree with our perception, we need an
alternate proposal.

So where can we go from here? We argue above that surface based
grouping hypothesis should not be rejected. So we may just
discredit triangulation based grouping hypothesis and
contemplate other means of interpolation? Although our
experiments rejected the Delaunay triangulation based
interpolation hypothesis, we are hesitant to discount
triangulation entirely. After all, realistic rendering of a 3D
world has been possible using triangles alone, and
computational efficiency and algorithmic simplicity associated
with triangulation are too attractive to abandon easily. At
this stage of our investigation, computational consideration is
important as we want to be able to generate visual stimuli and
test our proposals easily, consistently, and repeatedly.

By examining $\Omega_A$ and $\Omega_T$ representations of
shapes, we feel that the main sources of distraction in them
are triangles whose shapes are acutely different from others.
Some triangles are quite large and some are thin. Our percept
can be drawn to these salient triangles and gain sense of
disjointness with a large triangle and coherency with a cluster
of thin triangles. Thus, maybe a way to group isolated dots in
a computationally efficient and perceptually agreeable manner
is to use a more homogeneous set of triangles rather than
Delaunay ones. One way to achieve such interpolation is to
sub-divide a large triangle or a set of thin triangles by
introducing new internal points. The approach is common for
image synthesis of a parametric surface.

There are a number of advantages with this approach. First, by
using triangles as the construction blocks, we can inherit some
of advantages of the Delaunay based grouping algorithm,
including its computational efficiency and simplicity,
non-dependence on orientation and curvature information, and
capability to form illusory contours. We can also use the
Delaunay to draw an initial configuration, from which a more
homogeneous configuration can be derived. Thus, implementation
of the approach seems quite feasible. Second, we can achieve an
arbitrary level of homogeneousness by applying the subdivision
repeatedly. Thus, a smoother interpolation with more elaborate
construction blocks can be approximated by the approach. Third,
we may be able to use the representation to decompose a
non-convex shape into convex parts. Then, we will be able to
handle multiple objects, in which each object forms a convex
shape but joined together into a non-convex outline.
Polynomial-time algorithms have been known to decompose a
non-convex polygon into a set of convex polygons
\cite{Chazelle:1979}. The algorithms are not applicable here as
there is a great amount of ambiguity in forming an outer
polygon from a set of cues. (See Figures
\ref{fig:contourResults} and \ref{fig:surfaceResults}.) Another
approach is to first compute the distance transform on an image
lattice where each distance value tells the distance to the
nearest cue \cite{Kubota:MIA2011}. Local maxima of the distance
map gives convex cores, which can be used to group lattice
points into a collection of convex sets. By having a uniform
triangulation, we avail ourselves with a non-uniform lattice,
which can be used to derive a distance map, which in turn can
be used to separate multiple objects. Fourth, by setting a
threshold value for filtering local maxima and decreasing it
from a higher value to a lower one, we can construct a
scale-space representation of a set of convex parts. Such
representation has been shown effective in encoding both global
and local information for object recognition.
\cite{Witkin:ScaleSpaceA,Mokhtarian:Shape}.

\section{Conclusion}
The results of the cognitive study showed that shape
recognition performance was dependent on the representations
among dots, Delaunay triangles, and Delaunay triangles inside
the shape. The results reject out initial conjecture that
non-directional cues were grouped via Delaunay triangulation.
The results of the computational study showed a number of
advantages for the triangulation based grouping and recognition
over contour based ones.

Our final thought on the problem of grouping non-directional
visual cues is to further explore non-Delaunay triangulation,
in particular adaptive triangulation that yields a
representation comprised of visually similar triangles. We will
be able to obtain such triangulation from Delaunay one by
repetitive sub-division. We can then conduct cognitive
experiments similar to the ones presented in this paper using
the new triangulation.

\section*{Acknowledgement}
This work is supported by NSF grant CCF-1117439. J.R. thanks
Dr. Ernest Greene at University of Southern California for his
valuable inputs and feedbacks.

\end{document}